\documentclass{article}
\usepackage{microtype}
\usepackage{graphicx}
\usepackage{subfigure}
\usepackage{booktabs} %
\usepackage{multirow}
\usepackage{amsmath}
\usepackage{hyperref} 
\usepackage{cleveref}

    \usepackage[preprint]{neurips_2022}

\usepackage[utf8]{inputenc} %
\usepackage[T1]{fontenc}    %
\usepackage{url}            %
\usepackage{booktabs}       %
\usepackage{amsfonts}       %
\usepackage{nicefrac}       %
\usepackage{microtype}      %
\usepackage{xcolor}         %

\title{Set Features for Fine-grained Anomaly Detection}

\author{%
\qquad Niv Cohen \qquad  Issar Tzachor \qquad Yedid Hoshen \\
School of Computer Science and Engineering \\
The Hebrew University of Jerusalem, Israel \\ 
  \texttt{nivc@cs.huji.ac.il} \\
}

\begin{document}

\maketitle

\begin{abstract}

Fine-grained anomaly detection has recently been dominated by segmentation-based approaches. These approaches first classify each element of the sample (e.g., image patch) as normal or anomalous and then classify the entire sample as anomalous if it contains anomalous elements. However, such approaches do not extend to scenarios where the anomalies are expressed by an unusual combination of normal elements. In this paper, we overcome this limitation by proposing set features that model each sample by the distribution of its elements. We compute the anomaly score of each sample using a simple density estimation method. Our simple-to-implement approach\footnote{The code is available on github under: \href{https://github.com/NivC/SINBAD}{https://github.com/NivC/SINBAD}.} outperforms the state-of-the-art in image-level logical anomaly detection ($+3.4 \%$) and sequence-level time series anomaly detection ($+2.4 \%$). 
\end{abstract}

\section{Introduction}

\begin{figure*}
    
  \centering
 \includegraphics[width=0.99\linewidth]{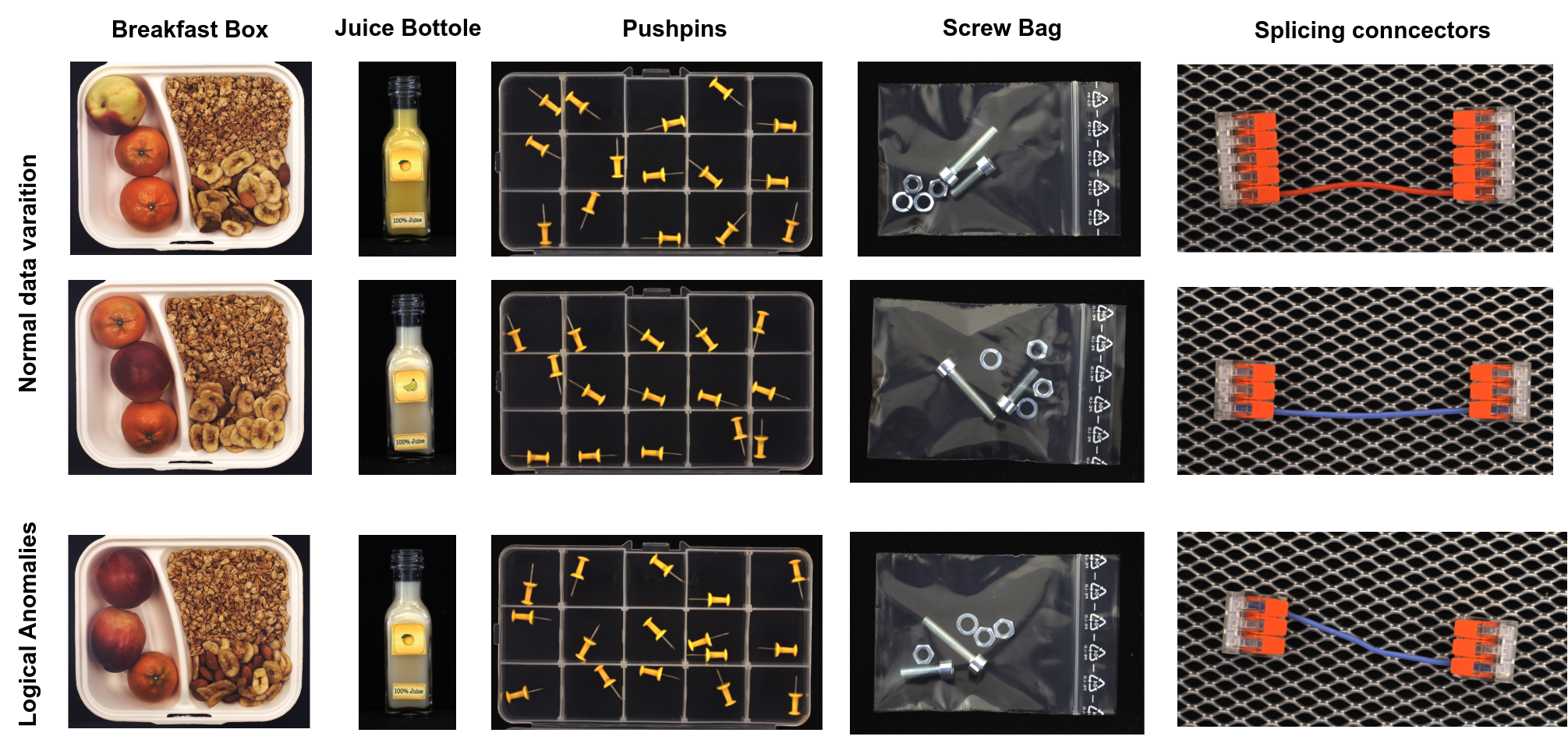} \\  
  \caption{  In logical anomalies, each image element (e.g., patch) may be normal even when their combination is anomalous. These cases are challenging as the variation among the normal data can be large while anomalies are fine-grained (e.g., swapping a bolt and a washer in the \textit{screw bag} class). }
  \label{fig:screw_logical}
  \vspace{5pt}
\end{figure*}

Anomaly detection aims to automatically identify samples that exhibit unexpected behavior. In fine-grained anomaly detection, such as detecting faults in industrial images or irregularities in time series, anomalies are quite subtle. For example, let us consider an image of a bag containing screws, nuts and washers (Fig.\ref{fig:screw_logical}). There are two ways in which a sample can be anomalous: (i) one or more of the elements in the sample are anomalous. E.g., a broken screw. (ii) the elements are normal but appear in an anomalous combination. E.g., one of the washers might be replaced with a nut.

In recent years, remarkable progress was made in detecting samples featuring anomalous elements. The usual procedure is:
First, we perform anomaly segmentation by detecting which (if any) of the elements of the sample are anomalous. This can be performed by a variety of methods, in particular, using density estimation methods \cite{cohen2020sub,defard2021padim,roth2022towards}. Given an anomaly segmentation map, we compute the sample-wise anomaly score as the number of anomalous elements, or the abnormality level of the most anomalous element. If the anomaly score exceeds a threshold, the entire sample is denoted as an anomaly. We denote this paradigm: detection-by-segmentation.

Here, we tackle the more challenging case of detecting anomalies consisting of an unusual combination of normal elements. For example, consider the case where normal images contain two washers and two nuts but anomalous images may contain one washer and three nuts. As each of the elements (nuts or screws) occur in natural images, simple detection-by-segmentation will not work. Instead, a more holistic understanding of the image is required. While simple global representations, e.g., taking the average of the representations of all elements, might work in some cases, the result is typically too coarse to detect fine-grained anomalies.  

We propose to detect anomalies consisting of unusual combinations of normal elements using set representations. The key assumption behind our method is that the distribution of elements in a sample is more correlated with it being anomalous than the ordering of the elements. Each sample is therefore modeled as an orderless set of elements. The elements are represented using feature embedding, e.g., a deep representation extracted by a pre-trained neural network or handcrafted features. To describe this set of features we count the percentage of elements falling in different histograms bins. We compute a histogram for each dimension of the feature space. The bins from all the histograms are concatenated together, forming our set representation. Finally, we score anomalies using density estimation on this set representation.

Our method, \textit{SINBAD} (\textit{Set} \textit{IN}spection
\textit{B}ased \textit{A}omalies \textit{D}etection) is evaluated on two diverse tasks. The first task is image-level logical anomaly detection on the MVTec-LOCO datasets. Our method outperforms more complex state-of-the-art methods, while not requiring any training. We also evaluate our method on series-level time series anomaly detection. Our approach outperforms all current methods while not using augmentation or training. %

We make the following contribution:

\begin{itemize}
\item Identifying set representation as key for detecting anomalies consisting entirely of normal elements.
\item An effective approach for measuring the distance between samples treated as sets of their local elements.
\item State-of-the-art results for sample-wise logical anomaly detection (MVTec-LOCO) and time series datasets.
\end{itemize}

\section{Previous work}

\textbf{Image Anomaly Detection.}
The field of anomaly detection has been researched for several decades. A comprehensive review can be found in \cite{ruff2021unifying}. %
Early approaches (\cite{glodek2013ensemble,latecki2007outlier,eskin2002geometric}) used handcrafted representations and aimed at detecting images with different coarse-grained objects (e.g., cats vs. dogs). Deep learning has provided a significant improvement on such benchmarks \cite{larsson2016learning, ruff2018deep,golan2018deep,hendrycks2019using,ruff2019deep,perera2019learning,mkd,csi}. As density estimation methods utilizing pre-trained deep representation have made significant steps towards the supervised performance on such benchmarks \cite{deecke2021transfer,cohen2022transformaly,panda,mean_shifted,reiss2022anomaly}, much research is now directed at other challenges \cite{reiss2022anomaly}. Such challenges include detecting anomalous image parts which are small and fine-grained \cite{cohen2020sub,li2021cutpaste,defard2021padim,roth2022towards,horwitz2022empirical}. The recent progress in anomaly detection and segmentation methods for this setting has been enabled by the introduction of appropriate datasets \cite{bergmann2019mvtec,bergmann2021mvtec,carrera2016defect,jezek2021deep}. The dominant paradigm in state-of-the-art methods \cite{roth2022towards} is to detect fine-grained anomalous images by first segmenting highly anomalous patches and then scoring the entire image based on these segmentation maps. Recently, the MVTec-LOCO dataset\cite{bergmann2022beyond} has put the spotlight on fine-grained anomalies which cannot be identified using single patches, but only when examining the connection between different (otherwise normal) elements in an image. Here, we will focus on detecting such \textit{logical} anomalies.

\textbf{Time series Anomaly detection.} The task of anomaly detection in time series has been studied over several decades \citep{blazquez2021review}. In this paper, we are concerned with anomaly detection of entire sequences, i.e. cases where an entire signal may be abnormal. Traditional approaches for this task include generic anomaly detection approaches such as $k$ nearest neighbors ($k$NN) based methods e.g. vanilla $k$NN \citep{eskin2002geometric} and Local Outlier Factor (LOF) \citep{breunig2000lof}, Tree-based methods \citep{liu2008isolation}, One-class classification methods \citep{tax2004support} and SVDD \citep{scholkopf1999support}. Some traditional methods such as auto-regressive methods are particular to time series anomaly detection \citep{rousseeuw2005robust}. With the advent of deep learning, the traditional approaches were augmented with deep-learned features. Deep one-class classification methods include DeepSVDD \citep{ruff2018deep} and DROCC \citep{goyal2020drocc}. Deep auto-regressive methods include RNN-based prediction and auto-encoding methods \citep{bontemps2016collective, malhotra2016lstm}. In addition, some deep learning anomaly detection approaches were proposed that are conceptually different from traditional approaches. These methods are based on the premise that classifiers trained on normal data will struggle to generalize to anomalous data. These approaches were originally developed for image anomaly detection \citep{golan2018deep} but have been extended to tabular and time series data \citep{bergman2020classification, qiu2021neural}.

\textbf{Discretized Projections.} Discretized projections of multivariate data have been used in many previous works. Locally sensitive hashing \cite{dasgupta2011fast} uses random projection and subsequent binary quantization as a hash for high-dimensional data. It was used to facilitate fast $k$ nearest neighbor search. Random projections transformation is also highly related to the Radon transform \cite{radon20051}. Kolouri et al. \citep{kolouri2015radon} used this representation as a building block in their set representation. HBOS \cite{goldstein2012histogram} performs anomaly detection by representing each dimension of multivariate data using a histogram of discretized variables and subsequent density estimation. LODA \cite{pevny2016loda} extends this work, by first projecting the data using a random projection matrix (followed by discretization). We differ from LODA in the use of a different density estimator and in using sets of multiple elements rather than single sample descriptions. Rocket and mini-rocket \cite{dempster2020rocket, dempster2021minirocket} represent time series for classification using the averages of their window projection.

\section{Set Features for Anomaly Detection}
\label{sec:method}

\subsection{A Set is More Than the Sum of Its Parts }
\label{subsec:motivation}
Detecting logical image anomalies, or collective time series anomalies, requires understanding how the different parts of each sample interact with one another. As a motivating example let us consider the \textit{screw bag} class from the MVTec-LOCO dataset (Fig.~\ref{fig:screw_logical}). Each normal sample in this class contains two screws (of different lengths), two nuts, and two washers. Anomalies may occur for example when one screw is missing, or when an additional nut replaces one of the washers. Detecting anomalies such as these requires a joint description of all elements within the sample since each local element could have come from a normal sample. 

\begin{figure}
  \centering
  \includegraphics[width=0.6\linewidth]{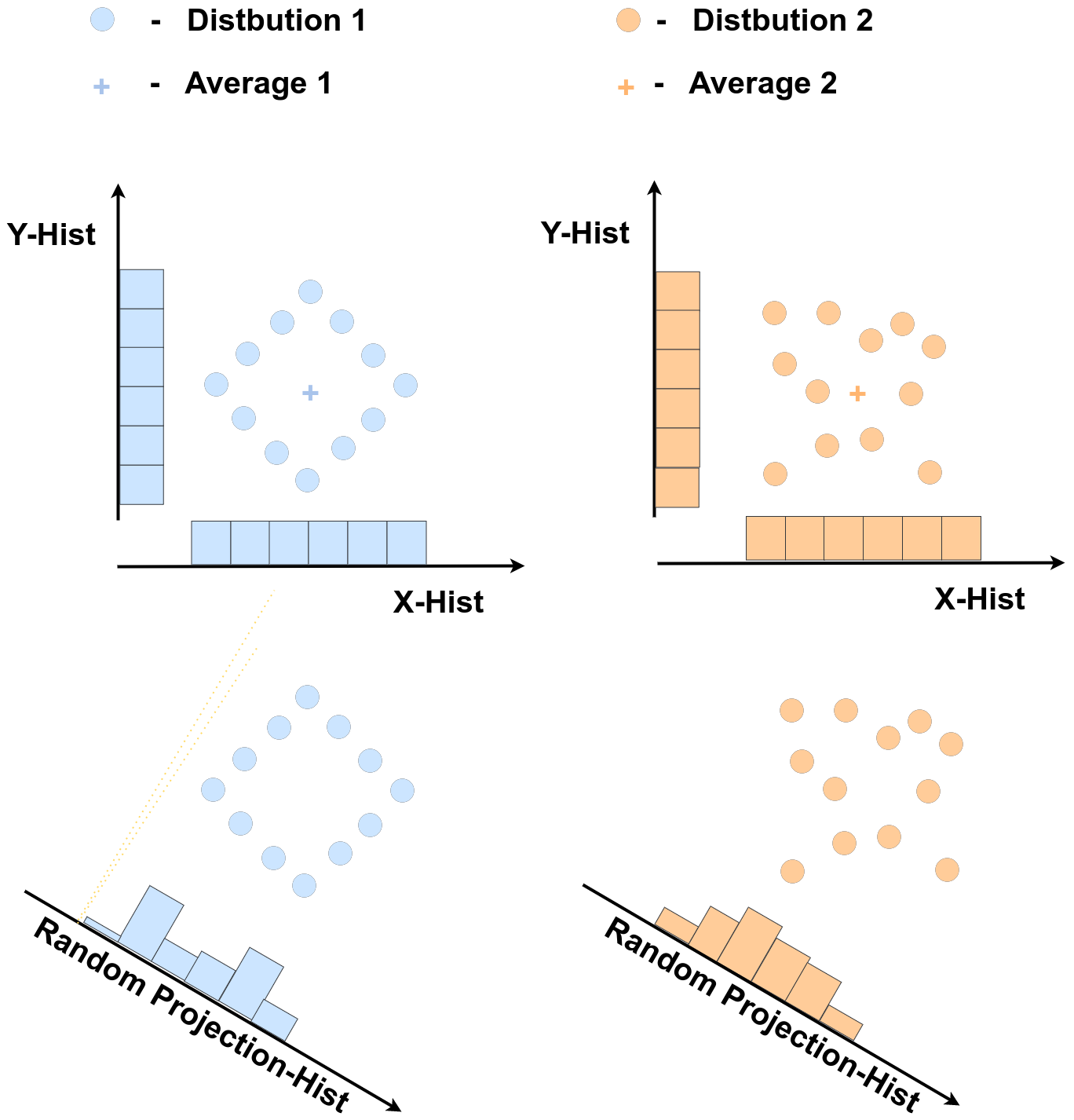}

  \caption{ Random projection histograms allow us to distinguish between sets where other methods could not. The two sets are similar in their averages and histograms along the original axes, but result in different histograms when projected along a random axis. }
  \label{fig:set_hists}
  \vspace{5pt}
\end{figure}

The typical way to aggregate element descriptor features %
is by average pooling. However, this is not always suitable for set anomaly detection. In supervised learning, average pooling is often built into architectures such as ResNet \cite{he2016deep} or DeepSets \cite{zaheer2017deep}, in order to aggregate local features. Therefore, deep features learnt with a supervised loss are already trained to be effective for pooling. However, for lower-level feature descriptors, such as the ones we use here, this may not be the case. As demonstrated in Fig.\ref{fig:set_hists}, the average of a set of features is far from a complete description of the set. This is especially true in anomaly detection, where density estimation approaches require more discriminative features than those needed for supervised learning \cite {reiss2022anomaly}. This means that even when an average pooled set of features worked for a supervised task, it might not work for anomaly detection.

Here, we wish to describe a set by modeling the distribution of its elements, ignoring the ordering between them. A naive way of doing so is by a discretized, volumetric representation, similarly to $3$D voxels for point clouds. Unfortunately, such approaches cannot scale to high dimensions, and more compact representations are required. %
In this work, we choose to represent sets using a collection of $1$D histograms. Each histogram represents the density of the elements of the set when projected along a particular direction. We provide an illustration of this idea in \Cref{fig:set_hists}.

Projecting a set along its original axes may not be discriminative enough. Histograms along the original axes correspond to 1D marginals, and may map distant elements to the same histogram bins (see \ref{fig:set_hists} for an illustration). On the other side, we can see at the bottom of the figure that when the set elements are first projected along another direction, the histograms of the two sets are distinct. This suggests a method for set description: first project each set along a shared random direction and then compute a 1D histogram for each set along this direction. We can obtain a more powerful descriptor by repeating this procedure with projections along multiple random directions.

\subsection{Preliminaries}
\label{subsec:prelim}

We are provided a training set $\mathcal{S}$ containing a set of $N_S$ samples $x_1,x_2..x_{N_S} \in \mathcal{S}$. All the samples at training time are known to be normal. At test time, we are presented with a new sample $\tilde{x}$. Our objective is to learn a model, which operates on each sample $\tilde{x}$ and outputs an anomaly score. Samples with anomaly scores higher than a predetermined threshold value are labeled as anomalies.

The unique aspect of our method is its treatment of each sample $x$ as consisting of a set of $N_E$ elements $x = [e_1,e_2..e_{N_E}]$. Examples of such elements include patches for images and temporal windows for time series. We assume the existence of a powerful feature extractor $F$ that maps each raw element $e$ into an element feature descriptor $f_e$. We will describe specific implementations of the feature extraction for two important applications: images and time series, in \cref{sec:applications}.

\subsection{Set Features by Histogram of Projections} 
\label{subsec:method_proj}

Motivated by the toy example in \cref{subsec:motivation}, we propose to model each set $x$ by computing a histogram of the values of its elements along each direction. As the given raw axes of the representation may mask out interesting degrees of variation, we perform a random projection prior to building the histograms.  

\textbf{Histogram descriptor.} As explained in \cref{subsec:motivation}, average pooling the features of all elements in the set may result in insufficiently informative representations. Instead, we describe the set using the histogram of values along each dimension. We note the set of the values of the $j$th feature in each element of each sample as $s[j] = \{f_1[j],f_2[j]..f_{N_S}[j]\}$. We compute the maximal and minimal values for sets $s[j]$ across all the samples, and divide the region between them into $K$ bins. %
We compute histograms $H_j$ for each of the $N_D$ dimensions and concatenate them into a single set descriptor $h$. The descriptor of each set therefore has a dimension of $N_D \cdot K$.    

\textbf{Projection.} As discussed before, not all projection directions are equally informative for describing the distributions of sets. %
In the general case, it is unknown which directions will be the most informative ones for capturing the difference between normal and anomalous sets. As we cannot tell the best projection directions in advance, we propose to randomly project the features. This ensures a low likelihood for catastrophically poor projection directions, such as those in the example in Fig.\ref{fig:set_hists}.

In practice, we generate a random projection matrix $P \in R^{(N_D,N_P)}$ by sampling values for each dimension from the Gaussian distribution $\mathbb{N}(0, 1)$. We project the features $f$ of each element of $x$, yielding projected features $f'$:

\begin{equation}
    f' = Pf
\end{equation}

We run the histogram descriptor procedure described above on the projected features. The final set descriptor $h_x$ therefore becomes the concatenation of $N_P$ histograms, resulting in a dimension of $N_P \cdot K$.

\begin{figure*}
  \centering
  \includegraphics[width=1\linewidth]{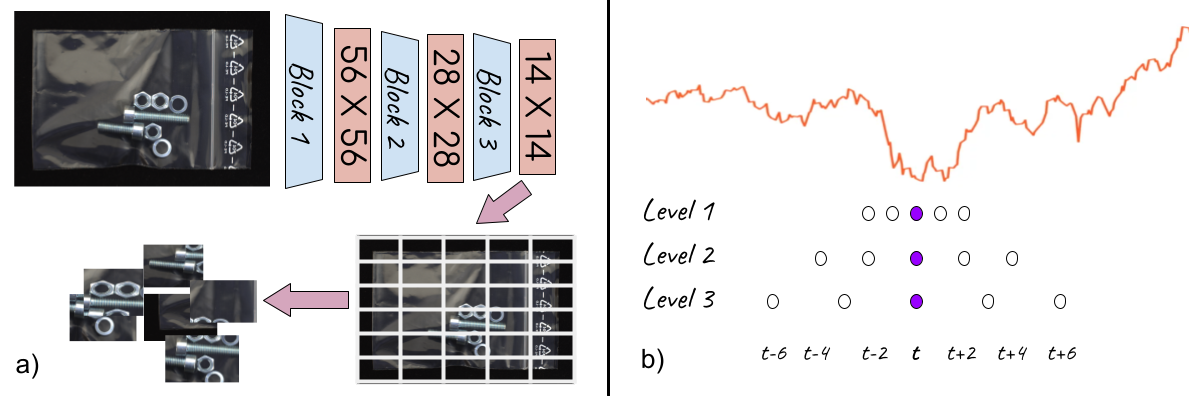}

  \caption{ For both image and time series samples we extract set elements of different granularity. In image samples (left), the sets of elements are extracted from different ResNet levels. For time series data (right), we take pyramids of windows at different strides around each time step (noted in blue circles).}
  \label{fig:elemnts}
  \vspace{5pt}
\end{figure*}

\subsection{Anomaly scoring}
\label{sec:anomaly_scoring}
We now wish to use the histogram features $h_x$ to detect anomalous samples. We estimate the probability density function (PDF) of the normal data features $\{h_x\} \in S$ using a Gaussian density estimator. We provide theoretical justification for the distributional assumption in App.~\ref{app:gaussian}.  We compute the mean $\mu$ and covariance $\Sigma$ of the descriptors of all sets. The estimated PDF is given by:
\begin{equation}
    p(h) = \mathcal{N}(h|\mu, \Sigma)
\end{equation}

As the anomaly score is correlated with the negative-likelihood of a sample, we define the anomaly score $a(h)$ as the negative log-likelihood. Ignoring constant terms, this becomes the Mahalanobis distance:

\begin{equation}
    a(h) = (h - \mu)^T \Sigma^{-1} (h - \mu)
\end{equation}

In practice, we found that the Mahalanobis distance to the $k$NN ($k$NN with whitening transformation) rather than to $\mu$ worked slightly better and is therefore used to compute our results.

\section{Application to Image and Time Series Anomaly Detection}
\label{sec:applications}

In this section, we apply our set description method for anomaly detection in image and time series data. 

\subsection{Images as Sets}
\label{subsec:images_sets}

Images can be seen as consisting of a set of elements of different levels of granularity. This ranges from pixels to small patches and low-level elements such as lines or corners, up to high-level elements such as objects. For anomaly detection, we typically do not know in advance the correct level of granularity for separating between normal and anomalous samples. This depends on the anomalies, which are unknown during training. Instead, we perform anomaly detection for different levels of granularity and combine the scores. These levels of granularity correspond to patches of different sizes. 

In practice, we use representations from intermediate blocks of a pre-trained ResNet \cite{he2016deep}. As a ResNet network simultaneously embeds many local patches of each image, we pass the image through the network encoder and extract our representations from the intermediate activations at the end of different residual blocks (see Fig.\ref{fig:elemnts}). We define each spatial location in the activation map as an element. Note that as different blocks have different resolutions, they yield different numbers of elements. We take the elements at the end of each residual block as our sets.

\subsection{Time Series as Sets}
\label{subsec:timeseries_sets}

Time series data can be viewed as a set of temporal windows. Similarly to images, it is generally not known in advance which temporal scale is relevant for detecting anomalies; i.e. the duration of windows which includes the semantic phenomenon. Inspired by \textit{Rocket} \cite{dempster2020rocket}, we define the basic elements of a time series as a collection of temporal window pyramids. Each pyramid contains $L$ windows. All the windows in a pyramid are  centered at the same time step, each containing $\tau$ samples (Fig.\ref{fig:elemnts}). The first level window includes $\tau$ elements with stride $1$, the second level window includes $\tau$ elements with stride $2$, etc. Such window pyramid is computed for each time step in the time series, and the entire series is represented as the set of its pyramid elements. More implementation details are described in Sec.\ref{app:imp_ts}.

\section{Results}

\begin{table*}[t]
\caption{Anomaly detection on MVTec-LOCO. ROC-AUC  ($\%$).  See Tab.\ref{tab:loco_anomalies_supp} for the full table.}
\centering
\small
\begin{tabular}{lcccccccccccc}
\toprule

		 &  f-AnoGAN &	MNAD	 & ST	& SPADE & PCore	& GCAD	& SINBAD \\ \midrule
\multirow{6}{*}{\rotatebox[origin=c]{90}{\scriptsize{\textbf{Logical Anomalies}}}} Breakfast box			&	69.4	&	59.9	&	68.9	&	81.8	&	77.7 & \underline{87.0}	&	\textbf{96.5}	$\pm$	\textbf{0.1}	\\			
\hspace{0.23cm} Juice bottle		&	82.4	&	70.5	&	82.9	&	91.9	& 83.7 &	\textbf{100.0}	&	\underline{96.6}	$\pm$	\underline{0.1}	\\			
\hspace{0.23cm} Pushpins		&	59.1	&	51.7	&	59.5	&	60.5	& 62.2	& \textbf{97.5}	&	\underline{83.4}	$\pm$	\underline{3.0}	\\			
\hspace{0.23cm} Screw bag		&	49.7	&	\underline{60.8}	&	55.5	&	46.8	& 55.3 &	56.0	&	\textbf{78.6}	$\pm$	\textbf{0.1}	\\			
\hspace{0.23cm}  Splicing connectors	&	68.8	&	57.6	&	65.4	&	73.8	& 63.3 &	\textbf{89.7}	&	\underline{89.3}	$\pm$	\underline{0.2}	\\			
\hspace{0.23cm} Avg. Logical	&	65.9	&	60.1	&	66.4	&	71.0	&  69.0 &	\underline{86.0}	&	\textbf{88.9}	$\pm$	\textbf{0.6}	\\			

 \midrule

\multirow{6}{*}{\rotatebox[origin=c]{90}{\scriptsize{\textbf{Structural Anoma.}}}} 
 Breakfast box		&	50.7	&	60.2	&	68.4	&	74.7	&	74.8 & \underline{80.9}	&	\textbf{87.5}	$\pm$	\textbf{0.1}	\\
\hspace{0.23cm} Juice bottle		&	77.8	&	84.1	&	\textbf{99.3}	&	84.9	& 86.7 &	\underline{98.9}	&	93.1	$\pm$	0.3	\\
\hspace{0.23cm} Pushpins	&	74.9	&	76.7	&	\textbf{90.3}	&	58.1	& \underline{77.6} &	74.9	 &	74.2	$\pm$	17.4	\\
\hspace{0.23cm} Screw bag	&	46.1	&	56.8	&	\underline{87.0}	&	59.8	& 86.6 &	70.5	&	\textbf{92.2}	$\pm$	\textbf{0.8}	\\
\hspace{0.23cm}  Splicing connectors	&	63.8	&	73.2	&	\textbf{96.8}	&	57.1	& 68.7 &	\underline{78.3}	&	76.7	$\pm$	0.2	\\
\hspace{0.23cm} Avg. Structural	&		62.7	&	70.2	&	\textbf{88.3}	&	66.9	&	78.9  & 80.7	&	\underline{84.7}	$\pm$	\underline{3.4}	\\
 \midrule

Avg. Total		&	64.3	&	65.1	&	77.4	&	68.9	&	74.0 & \underline{83.4}	&	\textbf{86.8} $\pm$ \textbf{1.8}	\\
 
\bottomrule
\end{tabular}
\label{tab:loco_anomalies}
\end{table*}

\subsection{Logical Anomaly Detection Results}

\textbf{Logical Anomalies Dataset.} We use the recently published MVTec-LOCO dataset \cite{bergmann2022beyond} to evaluate our method's ability to detect anomalies caused by unusual configurations of normal elements. This dataset features five different classes: \textit{breakfast box}, \textit{juice bottle}, \textit{pushpins}, \textit{screw bag} and \textit{splicing connector} (see Fig.\ref{fig:screw_logical}).  Each class includes: (i) a training set of normal samples exhibiting the normal variation of the class ($\sim 350$ samples). (ii) a validation set, containing another, smaller, set of normal samples ($\sim 60$ samples). (iii) a test set, containing normal samples, structural anomalies, and logical anomalies ($\sim 100$ samples each). 

The anomalies in each class are divided into \textit{structural anomalies} and \textit{logical anomalies}. Structural anomalies feature local defects, somewhat similar to previous datasets such as \cite{bergmann2019mvtec}. Conversely, logical anomalies may violate `logical' conditions expected from the normal data. As one example, an anomaly may include a different number of objects than the numbers expected from a normal sample (while all the featured object types exist in the normal class (Fig.\ref{fig:screw_logical})). Other types of logical anomalies in the dataset may include cases where distant parts of an image must correlate with one another. E.g., within the normal data, the color of one object may correlate with the length of another object. These correlations may break in an anomalous sample. 

\textbf{Metric.} Following the standard in image-level anomaly detection we report image level ROC-AUC metric. We report this metric individually for each anomaly type, for each data class.

\textbf{Baselines.} We compare to baseline methods used by the paper which presented the MVTec-LOCO dataset \cite{bergmann2022beyond}. Namely, we compare to \textit{Variational Model (VM)} \cite{steger2001similarity} - a handcrafted similarity measure designed for robustness to different conditions, \textit{MNAD} \cite{park2020learning} - an autoencoder with a memory module, \textit{f-AnoGAN} \cite{schlegl2017unsupervised} - a generative model trained for the reconstruction of anomaly free images, \textit{AE / VAE} \cite{sakurada2014anomaly} - an autoencoder / variational autoencoder, \textit{Student Teacher} (ST) \cite{bergmann2020uninformed} - a student network aimed to give better reconstruction for the normal data,  \textit{SPADE} \cite{cohen2020sub} - a density estimation method using a pyramid of deep ResNet features, \textit{PatchCore} (PCore) \cite{roth2022towards} - a state-of-the-art method for structural anomalies, improving SPADE scoring function, \textit{GCAD} \cite{bergmann2022beyond} - a reconstruction based method, based on both local and global deep ResNet features. %

\textbf{Results.} We report our results on image-level detection of logical anomalies and structural anomalies in Tab.\ref{tab:loco_anomalies}. Interestingly, we find complementary strengths between our approach and GCAD, a reconstruction-based approach by \cite{bergmann2022beyond}. Although  GCAD performed better on specific classes (e.g., \textit{pushpins}), our approach provides better results on average.  Most notably, our approach provides non-trivial anomaly detection capabilities on the \textit{screw bag} class, while baseline approaches are close to the random baseline. The rest of the compared approaches performed significantly worse on all logical anomaly classes, as they rely on the abnormality of single patches. 

Our approach also provides an improvement in the detection of structural anomalies in some classes. This is somewhat surprising, as one may assume that detection-by-segmentation approaches would perform well in these cases. One possible reason for that is the high variability of the normal data in some of the classes (e.g., \textit{breakfast box}, \textit{screw bag}, Fig.\ref{fig:screw_logical}). This high variability may induce false positive detections for detection-by-segmentation approaches. Taken together, while different methods provide complementary strengths, on average, our method provides state-of-the-art results on logical anomaly detection, and on the dataset as a whole.

\begin{table*}
\caption{Anomaly detection on the UEA datasets, average ROC-AUC ($\%$) over all classes. \newline See Tab.\ref{tab:realworld_supp} for the full table. $\sigma$ presented in Tab.~\ref{tab:realworld_errorbounds}}
\centering
\small
\begin{tabular}{lcccccccccccc}
\toprule

	&	OCSVM	&	IF	&		RNN	&	ED	&	DSVDD	& DAG		&	GOAD	&	DROCC	&	NeuTraL	&	Ours	\\ \midrule
EPSY	&	61.1	&	67.7		&	80.4	&	82.6	&	57.6	& 72.2  &	76.7	&	85.8	&	92.6	&	\textbf{98.1}	\\
NAT	&	86.0	&	85.4		&	89.5	&	91.5	&	88.6 & 78.9	& 	87.1	&	87.2	&	94.5	&	\textbf{96.1}	\\
SAD	&	95.3	&	88.2	&	81.5	&	93.1	&	86.0 & 80.9	& 	94.7	&	85.8	&	\textbf{98.9}	&	97.8	\\
CT	&	97.4	&	94.3		&	96.3	&	79.0	&	95.7 & 89.8	& 	97.7	&	95.3	&	99.3	&	\textbf{99.7}	\\
RS	&	70.0	&	69.3		&	84.7	&	65.4	&	77.4 & 51.0	& 	79.9	&	80.0	&	86.5	& \textbf{92.3}		\\ \midrule
Avg.	&	82.0	&	81.0	&	86.5	&	82.3	&	81.1 & 74.6 	&	87.2	&	86.8	&	94.4	&	\textbf{96.8}	\\

\bottomrule
\end{tabular}
\label{tab:realworld}
\end{table*}

\subsection{Time series anomalies detection results}

\label{subsec:time_series}

\textbf{Time series dataset.} We compared on the five datasets used in NeurTraL-AD \cite{qiu2021neural}: \textit{RacketSports (RS).} Accelerometer and gyroscope recording of players playing four different racket sports. Each sport is designated as a different class. \textit{Epilepsy (EPSY).} Accelerometer recording of healthy actors simulating four different activity classes, one of them being an epileptic shock. \textit{Naval air training and operating procedures standardization (NAT).} Positions of sensors mounted on different body parts of a person performing activities. There are six different activity classes in the dataset. \textit{Character trajectories (CT).} Velocity trajectories of a pen on a WACOM tablet. There are $20$ different characters in this dataset.  \textit{Spoken Arabic Digits (SAD).} MFCC features of ten Arabic digits spoken by $88$ different speakers.

\textbf{Metric.} Following \citet{qiu2021neural}, we use the series-level ROC-AUC metric.

\textbf{Baselines.} We compare the results of several baseline methods reported by \citet{qiu2021neural}. The methods cover the following paradigms: \textit{One-class classification}: One-class SVM (OC-SVM), and its deep versions DeepSVDD  (``DSVDD'') \cite{ruff2018deep}, and the recently published DROCC \cite{goyal2020drocc}. 
\textit{Tree-based detectors}: Isolation Forest (IF) \cite{liu2008isolation}. \textit{Density estimation}: LOF, a specialized version of nearest neighbor anomaly detection \cite{breunig2000lof}.  DAGMM (``DAG'') \cite{zong2018deep}: density estimation in an auto-encoder latent space \textit{Auto-regressive methods} - RNN and LSTM-ED (``ED'') - deep neural network-based version of auto-regressive prediction models \cite{malhotra2016lstm}. %
\textit{Transformation prediction} - GOAD \cite{bergman2020classification} and NeuTraL-AD \cite{qiu2021neural} are based on transformation prediction, and are adaptations of RotNet-based approaches (such as GEOM \cite{golan2018deep}).

\textbf{Results.} Our results are presented in Tab.~\ref{tab:realworld}. We can observe that different baseline approaches are effective for different datasets. $k$NN-based LOF is highly effective for SAD which is a large dataset but achieves worse results for EPSY. Auto-regressive approaches achieve strong results on CT. Transformation-prediction approaches, GOAD and NeuTraL achieve the best performance of all the baselines. The learned transformations of NeuTraL achieved better results than the random transformations of GOAD.

Our method achieves the best overall results both on average and individually on all datasets apart from SAD (where it is comparable but a little lower than NeuTraL). Note that differently from NeuTraL, our method is far simpler, does not use deep neural networks and is very fast to train and evaluate. It also has fewer hyperparameters.

\subsection{Implementation Details} 
\label{sec:implementation_details}

We provide here the main implementation details for our image anomaly detection application.
Further implementation details for the image application can be found in
App.\ref{app:imp_image}. Implementation details for the time series experiments can be found in App.\ref{app:imp_ts}.

\textbf{ResNet levels.} We use the representations from the third and fourth blocks of a \textit{WideResNet50$\times$2} (resulting in sets size $7\times7$ and $14\times14$ elements, respectively). We also use all the raw pixels in the image as an additional set (resized to $224\times224$ elements). 

The total anomaly score is the average of the anomaly scores obtained for the set of 3rd ResNet block features, the set of 4th ResNet block features, and the set of raw pixels. The average anomaly score is weighted by the following factors $(1,1,0.1)$ respectively (see App.\ref{app:loco_robust} for our robustness to the choice of weighting factor).

\textbf{Multiple crops for image anomaly detection.} Describing the entire image as a single set might sometime lose discriminative power when the anomalies are localized. To mitigate this issue, we can treat only a part of an image as our entire set. To do so, we crop the image to a factor of $c$, and compare the elements taken only from these crops. We compute an anomaly score for each crop factor and for each center location. We then average over the anomaly scores of the different crop center locations for the same crop factor $c$. Finally, for each ResNet level (described above), we average the anomaly scores over the different crop ratios $c$. We use crop ratios of $\{1.0,0.7,0.5,0.33\}$. The different center locations are taken with a stride of $0.25$ of the entire image.

\textbf{Parameters.} For the image experiments, we use histograms of $K=5$ bins and $r=1000$ projections. For the raw-pixels layer, we used a projection dimension of $r=10$ and no whitening due to its low number of channels. To avoid high variance between runs, we averaged $16$ different repetitions for the raw-pixel scoring. We use $k=1$ for the $k$NN density estimation.

\subsection{Ablations} 
\label{sec:ablation}

Here, we present ablations for the image logical AD methods. For time series ablations, see \cref{app:ablations}.

\textbf{Using individual ResNet levels.}
In Tab.\ref{tab:abl_image} we report the results of our method when different components of our multi-level ResNet ensemble are removed. We report the results using only the representation from the third or fourth ResNet block (``Only 3 / 4''). We also report the results of using both ResNet blocks, but without the raw-pixels level (``No Pixels"). 
While specific variants may outperform on specific classes, our combination outperforms on average. 

\textbf{No multiple crops ablation.} We also report our results without the multiple crops ensemble (described in Sec.\ref{sec:implementation_details}). In this ablation we feed only the entire image for the set extraction stage (``Only full''). As expected, using multiple crops field is beneficial for classes where small components are relatively important.

\textbf{Simple averaging.}
We compare to a simple averaging of the set features (Fig. \ref{fig:set_hists}), ablating our entire set-features approach. This yields a significantly worse performance (Tab.\ref{tab:abl_image_no_raw} ``Sim. Avg.'').

\textbf{No random projection.} We also ablate our use of random projections as described in Sec.\ref{subsec:method_proj}. We replace the random histograms with similar histograms using the raw given features. The advantage of our approach can be seen in Tab.\ref{tab:abl_image_no_raw} (``No Proj.'').

\textbf{No whitening.}
Finally, we also ablate our Gaussian model of the set features. Our method uses the collection of all the normal samples to calculate the covariance of the set features of the normal sample and modify our set distance measure as described in Sec.\ref{sec:anomaly_scoring}. Also here our approach outperforms, highlighting the benefits of describing the normal data as a collection of sets (Tab.\ref{tab:abl_image_no_raw} ``No Whit.'').

\begin{table}
\caption{Ablation for logical image AD. ROC-AUC  ($\%$) .}
\centering
\small
\begin{tabular}{lcccccccccccc}
\toprule

	&	Only 3	&	Only 4	&	No pixels	&	Only full	& Ours 	\\
 \toprule

Breakfa.	&	93.9	&	95.6	&	95.7	&	\textbf{97.1}	&	96.5	\\
Juice bo.	&	93.3	&	\textbf{97.4}	&	96.4	&	96.7	&	96.6	\\
Pushpins	&	78.0	&	66.8	&	73.4	&	77.5	&	\textbf{83.4}	\\
Screw b.	&	\textbf{79.5}	&	71.0	&	77.2	&	76.6	&	78.6	\\
Splicing.	&	85.3	&	85.0	&	86.3	&	\textbf{91.3}	&	89.3	\\
\midrule											
Average	&	86.0	&	83.2	&	85.8	&	87.8	&	\textbf{88.9}	\\

\bottomrule
\end{tabular}
\label{tab:abl_image}
\end{table}

\begin{table}
\caption{Ablation for logical image AD. Compared using no multiple crops and no  raw-pixels level. ROC-AUC  ($\%$).}
\centering
\small
\begin{tabular}{lcccccccccccc}
\toprule

	&	 Sim. Avg.	&	No Proj.	&	No Whit.	&	Ours	\\
  \toprule
Breakfa.	&	88.4	&	91.2	&	94.2	&	\textbf{95.1}	\\
Juice bo.	&	95.7	&	94.7	&	93.4	&	\textbf{95.2}	\\
Pushpins	&	64.1	&	68.9	&	69.6	&	\textbf{73.4}	\\
Screw b.	&	56.5	&	62.3	&	66.4	&	\textbf{70.2}	\\
Splicing.	&	87.7	&	\textbf{89.2}	&	88.0	&	86.7	\\
\midrule									
Average	&	78.5	&	81.3	&	82.3	&	\textbf{84.1}	\\

\bottomrule
\end{tabular}
\label{tab:abl_image_no_raw}
\end{table}

\section{Discussion} 
\label{sec:discussion}

\textbf{Complementary strength of density estimation and reconstruction based approaches for logical  anomaly detection.} As our method and GCAD \cite{bergmann2022beyond}, a reconstruction based approach, exhibit complementary strengths, it is a natural direction to try and use them together. A practical way to take advantage of both approaches would be ensembling. A better understanding of the reasons for each method's different performance across classes is likely to lead to the development of better approaches, combining the strengths of each method.

\textbf{Is our set descriptor approach beneficial for detecting structural image  anomalies?} While our method slightly lags behind the top segmentation-by-detection approach on structural anomalies, it achieves the top performance on specific classes. We hypothesize this may be due to the high variation among the normal samples in these classes. In this case too, future research may allow the construction of better detectors, enjoying the combined strength of both approaches.

\textbf{Incorporating deep features for time series data.} We demonstrated that our method is able to outperform the state-of-the-art in time series anomaly detection without using deep neural networks. While this is an interesting and surprising result, we believe that deep features will be incorporated into similar approaches in the future. One direction for doing this is replacing the window projection features with a suitable deep representation, while keeping the averaging and Gaussian modeling steps unchanged. 

\textbf{Relation to previous methods and optimal transport.} Our method is related to several previous methods. HBOS \citep{goldstein2012histogram} and LODA \citep{pevny2016loda} also used similar projection features for anomaly detection. Yet, these methods perform histogram-based density estimation by ignoring the dependency across projections. As they can only be applied to a single element (time-window), they do not achieve competitive performance for time series AD. Rocket/mini-rocket \citep{dempster2020rocket, dempster2021minirocket} also average projection features across windows but do not tackle anomaly detection nor do they apply to image data. Finally, there is a subtle connection to Radon transform \citep{kolouri2015radon} and sliced Wasserstein distance in $L_1$ (SWD) based methods \citep{bonneel2015sliced} which also use similar projection and histogram features. %

\section{Limitations} 

\textbf{Element-level anomaly detection.} Our method focuses on sample-level time series and image-level anomaly detection. In some applications, a user may also want a segmentation of the most anomalous elements of each sample. We note that for logical anomalies, this is often not very well defined. E.g., when we have an image with $3$ nuts as opposed to the normal $2$, each of them may be considered anomalous. To provide element-level information, our method can be combined with current segmentation approaches by incorporating the knowledge of a global anomaly (e.g., removing false positive segmentation if an image is normal). However, directly applying our set features for anomaly segmentation is left for future research. 

\textbf{Pre-trained features.} Similarly to the other top-performing approaches, our approach for image anomaly detection relies on pre-trained features. While the use of pre-trained features for anomaly detection in images is standard, it has failure modes. There are a handful of datasets where ImageNet pretraining is known to fail \cite{yousef2023no}. 

\textbf{Class-specific performance.} While our method outperforms on average, in some classes we do not perform as well compared to baseline approaches. A better understanding of the cases when our method fails would be beneficial for deploying it in practice.

\textbf{Non-IID elements.} Our method uses a Gaussian anomaly scoring function. By the central limit theorem, this is justifiable when the elements are IID (see \cref{app:gaussian}). However, this is not precisely true for either of our settings as elements have a strong overlap. In general, we find the Mahalanobis scoring function is highly effective, and the combination with $k$NN relaxes the Gaussian assumption. A more careful analysis is left for future work.

\section{Conclusion}

We presented a method for detecting anomalies caused by unusual combinations of normal elements. We introduce set features dedicated to capturing such phenomena, and demonstrate their applicability for images and time series. Extensive experiments established the strong performance of our method.

\section{Acknowledgement}
Niv Cohen was funded by Israeli Science Foundation and the Hebrew University Data Science grants (CIDR). We thank Paul Bergmann for kindly sharing numerical results for many of the methods compared on the MVTec-LOCO dataset.

\bibliography{example_paper}
\bibliographystyle{icml2023}

\clearpage

\appendix

 \title{Appendix}

\section{Full Results Tables}
\label{app:full_results}
The full table image logical anomaly detection experiments can be found in Tab.\ref{tab:loco_anomalies_supp}.
The full table for the time series anomaly detection experiments can be found in Tab.\ref{tab:realworld_supp}.

\begin{table*}[b]
\caption{Anomaly detection on the MVTec-LOCO dataset. ROC-AUC  ($\%$).}
\centering
\small
\begin{tabular}{lcccccccccccc}
\toprule

	& VM &	AE &	VAE	 & f-AG &	MNAD	 \\ 
\midrule

\multirow{6}{*}{\rotatebox[origin=c]{90}{\scriptsize{\textbf{Logical Anomalies}}}} Breakfast box     &					70.3	&	58.0	&	47.3	&	69.4	&	59.9	 \\ 
\hspace{0.23cm} Juice bottle   	&	59.7	&	67.9	&	61.3	&	82.4	&	70.5   \\ 
\hspace{0.23cm} Pushpins   &	42.5	&	62.0	&	54.3	&	59.1	&	51.7   \\ 
\hspace{0.23cm} Screw bag    	&	45.3	&	46.8	&	47.0	&	49.7	&	\underline{60.8}  \\ 
\hspace{0.23cm}  Splicing connectors    &	64.9	&	56.2	&	59.4	&	68.8	&	57.6	  \\ 
\hspace{0.23cm} Avg. Logical   	&	56.5	&	58.2	&	53.8	&	65.9	&	60.1	   \\ 

\midrule

\multirow{6}{*}{\rotatebox[origin=c]{90}{\scriptsize{\textbf{Structural Anom.}}}}
Breakfast box    &				70.1	&	47.7	&	38.3	&	50.7	&	60.2  \\ 
\hspace{0.23cm} Juice bottle    &	69.4	&	62.6	&	57.3	&	77.8	&	84.1  \\ 
\hspace{0.23cm} Pushpins    	&	65.8	&	66.4	&	75.1	&	74.9	&	76.7  \\ 
\hspace{0.23cm} Screw bag  	&	37.7	&	41.5	&	49.0	&	46.1	&	56.8	    \\ 
\hspace{0.23cm}  Splicing connectors  &	51.6	&	64.8	&	54.6	&	63.8	&	73.2    \\ 
\hspace{0.23cm} Avg. Structural    &	58.9	&	56.6	&	54.8	&	62.7	&	70.2  \\ 

\midrule

Avg. Total    & 57.7	&	57.4	&	54.3	&	64.3	&	65.1  \\ 

 \midrule

		 & ST	& SPADE & PCore	& GCAD	& SINBAD \\ \midrule
\multirow{6}{*}{\rotatebox[origin=c]{90}{\scriptsize{\textbf{Logical Anomalies}}}} Breakfast box	&	68.9	&	81.8	&	77.7 & \underline{87.0}	&	\textbf{96.5}	$\pm$	\textbf{0.1}	\\			
\hspace{0.23cm} Juice bottle		&	82.9	&	91.9	& 83.7 &	\textbf{100.0}	&	\underline{96.6}	$\pm$	\underline{0.1}	\\			
\hspace{0.23cm} Pushpins			&	59.5	&	60.5	& 62.2	& \textbf{97.5}	&	\underline{83.4}	$\pm$	\underline{3.0}	\\			
\hspace{0.23cm} Screw bag		&	55.5	&	46.8	& 55.3 &	56.0	&	\textbf{78.6}	$\pm$	\textbf{0.1}	\\			
\hspace{0.23cm}  Splicing connectors		&	65.4	&	73.8	& 63.3 &	\textbf{89.7}	&	\underline{89.3}	$\pm$	\underline{0.2}	\\			
\hspace{0.23cm} Avg. Logical	&	66.4	&	71.0	&  69.0 &	\underline{86.0}	&	\textbf{88.9}	$\pm$	\textbf{0.6}	\\			

 \midrule

\multirow{6}{*}{\rotatebox[origin=c]{90}{\scriptsize{\textbf{Structural Anom.}}}} 
 Breakfast box		&	68.4	&	74.7	&	74.8 & \underline{80.9}	&	\textbf{87.5}	$\pm$	\textbf{0.1}	\\
\hspace{0.23cm} Juice bottle			&	\textbf{99.3}	&	84.9	& 86.7 &	\underline{98.9}	&	93.1	$\pm$	0.3	\\
\hspace{0.23cm} Pushpins		&	\textbf{90.3}	&	58.1	& \underline{77.6} &	74.9	 &	74.2	$\pm$	17.4	\\
\hspace{0.23cm} Screw bag	&	\underline{87.0}	&	59.8	& 86.6 &	70.5	&	\textbf{92.2}	$\pm$	\textbf{0.8}	\\
\hspace{0.23cm}  Splicing connectors			&	\textbf{96.8}	&	57.1	& 68.7 &	\underline{78.3}	&	76.7	$\pm$	0.2	\\
\hspace{0.23cm} Avg. Structural
	&	\textbf{88.3}	&	66.9	&	78.9  & 80.7	&	\underline{84.7}	$\pm$	\underline{3.4}	\\
 \midrule

Avg. Total		&	77.4	&	68.9	&	74.0 & \underline{83.4}	&	\textbf{86.8} $\pm$ \textbf{1.8}	\\
 
\bottomrule
\end{tabular}
\label{tab:loco_anomalies_supp}
\end{table*}

\begin{table*}
\caption{UEA datasets, average ROC-AUC ($\%$) over all classes. ($\sigma$ presented in Tab.~\ref{tab:realworld_errorbounds})}
\centering
\small
\begin{tabular}{lcccccccccccc}
\toprule

&	OCSVM	&	IF	&	LOF	&	RNN	&	ED	\\
\midrule

EPSY	& 61.1	&	67.7	&	56.1	&	80.4	&	82.6 \\
NAT	& 86.0	&	85.4	&	89.2	&	89.5	&	91.5 \\
SAD	&  95.3	&	88.2	&	98.3	&	81.5	&	93.1 \\
CT	& 97.4	&	94.3	&	97.8	&	96.3	&	79.0 \\
RS	&  70.0	&	69.3	&	57.4	&	84.7	&	65.4 \\
\midrule
Avg.	&  82.0	&	81.0	&	79.8	&	86.5	&	82.3	\\
\midrule
	&	DSVDD	& DAGMM		&	GOAD	&	DROCC	&	NeuTraL	&	Ours	\\ \midrule
EPSY	&	57.6	& 72.2  &	76.7	&	85.8	&	92.6	&	\textbf{98.1}	\\
NAT		&	88.6 & 78.9	& 	87.1	&	87.2	&	94.5	&	\textbf{96.1}	\\
SAD		&	86.0 & 80.9	& 	94.7	&	85.8	&	\textbf{98.9}	&	97.8	\\
CT		&	95.7 & 89.8	& 	97.7	&	95.3	&	99.3	&	\textbf{99.7}	\\
RS		&	77.4 & 51.0	& 	79.9	&	80.0	&	86.5	& \textbf{92.3}		\\ \midrule
Avg.	&		81.1 & 74.6 	&	87.2	&	86.8	&	94.4	&	\textbf{96.8}	\\

\bottomrule
\end{tabular}
\label{tab:realworld_supp}
\end{table*}

\section{UEA Results with Standard Errors}
\label{app:extended_uea_bounds}

We present an extended version of the UEA results including error bounds for our method and baselines that reported them. The difference between the methods is significantly larger than the standard error.

\begin{table*}[h]
\caption{UEA datasets, average ROC-AUC  ($\%$) over all classes including error bounds}
\centering
\small
\begin{tabular}{lccccccccccc}
\toprule

	&	OCSVM	&	IF	&	LOF	&	RNN			&	LSTM-ED			\\ \midrule						
EPSY	&	61.1	&	67.7	&	56.1	&	80.4	$\pm$	1.8	&	82.6	$\pm$	1.7	\\						
NAT	&	86	&	85.4	&	89.2	&	89.5	$\pm$	0.4	&	91.5	$\pm$	0.3	\\						
SAD	&	95.3	&	88.2	&	98.3	&	81.5	$\pm$	0.4	&	93.1	$\pm$	0.5	\\						
CT	&	97.4	&	94.3	&	97.8	&	96.3	$\pm$	0.2	&	79.0	$\pm$	1.1	\\						
RS	&	70	&	69.3	&	57.4	&	84.7	$\pm$	0.7	&	65.4	$\pm$	2.1	\\ \midrule						
Avg.	&	82.0	&	81.0	&	79.8	&	86.5			&	82.3			\\ \midrule						
	&	DeepSVDD	& DAGMM	&			GOAD			&	DROCC			&	NeuTraL			&	Ours			\\ \midrule
EPSY	&	57.6	$\pm$ 0.7	 & 72.2 $\pm$ 1.6	 &	76.7	$\pm$	0.4	&	85.8	$\pm$	2.1	&	92.6	$\pm$	1.7	&	\textbf{98.1}	$\pm$	0.3	\\
NAT	&	88.6	$\pm$	0.8	& 78.9 $\pm$ 3.2 &	87.1	$\pm$	1.1	&	87.2	$\pm$	1.4	&	94.5	$\pm$	0.8	&	\textbf{96.1}	$\pm$	0.1	\\
SAD	&	86.0	$\pm$	0.1	&  80.9 $\pm$ 1.2 & 	94.7	$\pm$	0.1	&	85.8	$\pm$	0.8	&	\textbf{98.9}	$\pm$	0.1	&	97.8	$\pm$	0.1	\\
CT	&	95.7	$\pm$	0.5 & 89.8 $\pm$ 0.7	&	97.7	$\pm$	0.1	&	95.3	$\pm$	0.3	&	99.3	$\pm$	0.1	&	\textbf{99.7}	$\pm$	0.1	\\
RS	&	77.4	$\pm$	0.7	& 51.0 $\pm$ 4.2	& 79.9	$\pm$	0.6	&	80.0	$\pm$	1.0	&	86.5	$\pm$	0.6	& \textbf{92.3}	$\pm$	0.3	\\ \midrule	
Avg.	&	81.1			& 74.6	& 87.2			&	86.8			&	94.4			&	\textbf{96.8	}		\\

\bottomrule
\end{tabular}
\label{tab:realworld_errorbounds}
\end{table*}

\section{Implementation Details}
\label{app:imp}

\textit{Histograms}. In practice, we use the cumulative histograms as our set features for both data modalities (of Sec.\ref{subsec:method_proj}).

\subsection{Image anomaly detection}
\label{app:imp_image}

 \textit{Preprocessing.} Before feeding each image sample to the pre-trained network we resize it to $224 \times 224$ and normalize it according to the standard ImageNet mean and variance.
 
 Considering that classes in this dataset are provided in different aspect ratios, and that similar objects may look different when resized to a square, we found it beneficial to pad each image with empty pixels. The padded images have a $1:1$ aspect ratio, and resizing them would not change the aspect ratio of the featured objects.

\textit{Software.} 
For the whitening of image features we use the \textit{ShrunkCovariance} function from the \textit{scikit-learn library} \cite{scikit-learn} with its default parameters. For $k$NN density estimation we use the \textit{faiss} library \cite{johnson2019billion}.

\textit{Computational resources.} The experiments were run on a single RTX2080-GT GPU.

\subsection{Time Series  anomaly detection}
\label{app:imp_ts}

\textit{Padding.} Prior to window extraction, the series $x$ is first right and left zero-padded by $\frac{\tau}{2}$ to form padded series $x'$. The first window $w_1$ is defined as the first $\tau$ observations in padded series $S'$, i.e. $w_1 = x'_1,x'_2..x'_{\tau}$. We further define windows at higher scales $W^s$, which include observations sampled with stride $c$. At scale $c$, the original series $x$ is right and left zero-padded by $\frac{c \cdot \tau}{2}$ to form padded series $S'^c$. 

\textit{UEA Experiments.} We used each time series as an individual training sample. We chose a kernel size of $9$, $100$ projection, $20$ quantiles, and a maximal number of levels of $10$. The results varied only slightly within a reasonable range of the hyperparameters e.g. using $5$, $10$, $15$ levels yielded an average ROCAUC of $97$, $96.8$, $96.8$ across the five UEA datasets.

\textit{Spoken Arabic Digits processing}  We follow the processing of the dataset as done by Qiu et al. \cite{qiu2021neural}.
 In private communications the authors explained that only sequences of lengths between 20 and 50 time steps were selected. The other time series were dropped.

\textit{Computational resources.} The experiments were run on a modest number of CPUs on a computing cluster. The baseline methods were run on a single RTX2080-GT GPU

\section{Logical Anomaly detection Robustness }
\label{app:loco_robust}
We check the robustness of our results for the parameter $\lambda$ - the weighting between the raw-pixels level anomaly score to the anomaly score derived from the ResNet features (Sec.\ref{sec:implementation_details}). As can be seen in Tab.\ref{tab:lambda_robust}, our results are robust to the choice of $\lambda$.

\begin{table}
\caption{Robustness to the choice of $\lambda$. Average Results on the Logical Anomalies classes. Average ROC-AUC  ($\%$).}
\centering
\begin{tabular}{lccccc}
\toprule

	$\lambda$ &	0.2	&	0.1 (Ours)		&	0.05	&	0.02	\\ \midrule

	&	88.8	&	\textbf{88.9}	&	88.7			&	88.5 \\

\bottomrule
\end{tabular}
\label{tab:lambda_robust}
\end{table}

\section{Time Series Ablations}
\label{app:ablations}

\begin{figure*}
  \centering
  \includegraphics[width=0.3\linewidth]{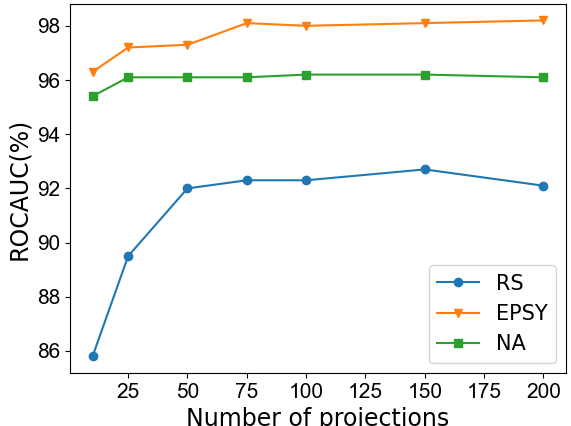}~~~
  \includegraphics[width=0.3\linewidth]{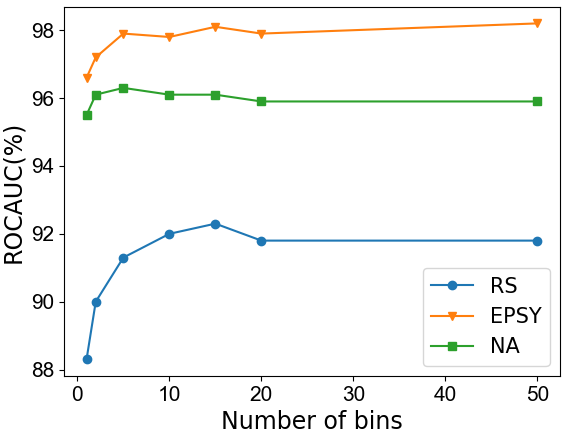}
  \caption{Ablation of accuracy vs. the number of projections (left) and the number of bins (right). }
  \label{fig:slice_bin}
  \vspace{5pt}
\end{figure*}

\textbf{Number of projections.} Using a high output dimension for projection matrix $P$ increases the expressively but also increases the computation cost. We investigate the effect of the number of projections on the final accuracy of our method. The results are provided in Fig.~\ref{fig:slice_bin}. We can observe that although a small number of projections hurts performance, even a moderate number of projections is sufficient. We found $100$ projections to be a good tradeoff between performance and runtime.

\textbf{Number of bins.} We compute the accuracy of our method as a function of the number of bins per projection. Our results ( Fig.~\ref{fig:slice_bin}) show that beyond a very small number of bins - larger numbers are not critical. We found $20$ bins to be sufficient in all our experiments.

\begin{table}
\caption{An ablation of projection sampling methods. ROC-AUC  ($\%$).}
\centering
\begin{tabular}{lccccc}
\toprule

	&	EPSY	&	RS	&	NA	&	CT	&	SAD	\\ \midrule

No whitening	&	62.1	&	70.9	&	93.6	&	98.5	&	78.8 \\
Whitening	&	\textbf{98.1}	&	\textbf{92.3}	&	\textbf{96.1}	&	\textbf{99.7}	&	\textbf{97.8}	\\ 

\bottomrule
\end{tabular}
\label{tab:ablation_cov}
\end{table}

\textbf{Effect of Gaussian density estimation.} Standard projection methods such as HBOS \cite{goldstein2012histogram} and LODA \cite{pevny2016loda} do not use a multivariate density estimator but instead estimate the density of each dimension independently. We compare using a full and per-variable density estimation in Tab.~\ref{tab:ablation_cov}. We can see that our approach achieves far better results, attesting to the importance of modeling the correlation between projections.

\begin{table}
\caption{An ablation of projection sampling methods. ROC-AUC  ($\%$).}
\centering
\begin{tabular}{lccccc}
\toprule
&	EPSY	&	RS	&	NA	&	CT	&	SAD	\\ \midrule
Id.	&	97.1	&	90.2	&	91.8	&	98.2	&	78.3	\\
PCA	&	\textbf{98.2}	&	91.6	&	95.8	&	\textbf{99.7}	&	96.7	\\
Rand	&	98.1	&	\textbf{92.3}	&	\textbf{96.1}	&	\textbf{99.7}	&	\textbf{97.8}	\\

\bottomrule
\end{tabular}
\label{tab:ablation_proj}
\end{table}

\textbf{Comparing projection sampling methods.} We compare three different projection selection procedures: (i) Gaussian: sampling the weights in $P$ from a random Normal Gaussian distribution (ii) Using an identity projection matrix: $P = I$ . (iii) PCA: selecting $P$ from the eigenvectors of the matrix containing all (raw) features of all training windows. PCA selects the projections with maximum variation but is computationally expensive. The results are presented in Tab.~\ref{tab:ablation_proj}. We find that the identity projection matrix under-performed the other approaches (as it provides no variable mixing). Surprisingly, we do not see a large difference between PCA and random projections.     

\section{Using the Central Limit Theorem for Set anomaly detection}
\label{app:gaussian}
We model the features of each window $f$ as a normal set as IID observations coming from a probability distribution function $p(f)$. The distribution function is \textit{not} assumed to be Gaussian. Using a Gaussian density estimator trained on the features of elements observed in training is unlikely to be effective for element-level anomaly detection (due to the non-Gaussian $p(f)$). 

An alternative formulation to the one presented in \cref{sec:method}, is that each feature $f$ is multiplied by projection matrix $P$, and then each dimension is discretized and mapped to a one-hot vector. This formulation therefore maps the representation of each element to a sparse binary vector. The mean of the representations of elements in the set recovers the normalized histogram descriptor precisely (therefore this formulation is equivalent to the one in \cref{sec:method}).
As the histogram is a mean of the set of elements, it has superior statistical properties. In particular, the Central Limit Theorem states that under some conditions the sample mean follows the Gaussian distribution regardless of the distribution of windows $p(f)$. While typically in anomaly detection only a single sample is presented at a time, the situation is different when treating samples as sets. Although the windows are often not IID, given a multitude of elements, an IID approximation may be approximately correct. This explains the high effectiveness of Gaussian density estimation in our formulation.

\end{document}